\documentclass[]{Liebert_Author_revised}
\usepackage{graphicx}
\usepackage{multicol}
\usepackage{multirow}

\usepackage{url}
\usepackage{color}
\usepackage{marvosym}
\usepackage{float}

\usepackage{caption}
\usepackage{subcaption}
\usepackage{cite}
\usepackage{amsmath,amssymb,amsfonts}
\usepackage{algorithmic}

\title{DANCE: \textbf{D}eep Learning-Assisted \textbf{A}nalysis of Protei\textbf{N} Sequences Using \textbf{C}haos \textbf{E}nhanced Kaleidoscopic Images} 

\author{Taslim Murad$^1$, Prakash Chourasia$^1$, Sarwan Ali$^2$, \\Imdad Ullah Khan$^3$,  and Murray Patterson$^1$\\
  {$^1$Department of Computer Science, Georgia State University}\\
  {Atlanta, GA 30303, USA}\\
  {$^2$Department of Neurology}\\
  {Columbia University, New York, NY, USA}\\
  {$^3$Department of Computer Science}\\
  {Lahore University of Management Sciences, Lahore, Pakistan}\\
E-mail: \{tmurad2,pchourasia1\}@student.gsu.edu, sa4559@columbia.edu, \\ 
  imdad.khan@lums.edu.pk, mpatterson30@gsu.edu
}

\date{}

\begin{document} 

\maketitle 


\begin{abstract}
Cancer is a complex disease characterized by uncontrolled cell growth, and requires accurate classification for effective treatment. T cell receptors (TCRs), crucial proteins in the immune system, play a key role in recognizing antigens. Advancements in sequencing technologies have facilitated comprehensive profiling of TCR repertoires, uncovering TCRs with potent anti-cancer activity and enabling TCR-based immunotherapies. Effectively analyzing these complex biomolecules requires representations that accurately capture both their structural and functional characteristics. 
T-cell protein sequences pose unique challenges due to their relatively smaller lengths than other biomolecules. Traditional vector-based embedding methods may encounter problems such as loss of information. Therefore, an image-based representation approach becomes a preferred choice for efficient embeddings, allowing for the preservation of essential details and enabling comprehensive analysis of T-cell protein sequences.
We propose to generate images from protein sequences using the idea of Chaos Game Representation (CGR). We design images using the Kaleidoscopic images approach. This \textbf{D}eep Learning-Assisted \textbf{A}nalysis of Protei\textbf{N} Sequences Using \textbf{C}haos \textbf{E}nhanced Kaleidoscopic Images (called DANCE) provides a unique way to visualize protein sequences by recursively applying chaos game rules around a central seed point. The resulting kaleidoscopic images exhibit symmetrical patterns that offer a visual representation of the protein sequences.
To investigate the effectiveness of this approach, we perform classification of the T cell receptors (TCR) protein sequences in terms of their respective target cancer cells, since TCRs are known for their immune response against cancer disease. The DANCE technique is used to turn the TCR sequences into pictures prior to classification. We employ deep-learning vision models to classify the generated images to obtain insights into the relationship between the visual patterns in the generated kaleidoscopic images and the underlying protein properties. By combining CGR-based image generation with deep learning classification, this study opens novel possibilities in protein analysis.
\end{abstract}

\section{Introduction}\label{sec_introduction}

Understanding and effectively analyzing T cell receptors (TCRs), crucial proteins involved in recognizing antigens associated with cancer, holds immense importance in cancer research and treatment~\citeauthor{li2020t}. Recent advancements in sequencing technologies have enabled comprehensive profiling of TCR repertoires, unveiling TCRs with potent anti-cancer activity and paving the way for TCR-based immunotherapies~\citeauthor{gohil2021applying}. However, the analysis of TCR protein sequences presents unique challenges. Compared to other biomolecules, TCR sequences are relatively shorter~\citeauthor{hou2016analysis}, making traditional vector-based embedding methods less suitable due to the potential loss of critical information.

Traditional embedding methods have been widely used for representing protein sequences~\citeauthor{ali2022pwm2vec}, aiming to capture their structural~\citeauthor{chen2021hydrogen} and functional characteristics~\citeauthor{du2020deepadd}. These methods typically involve transforming the protein sequences into fixed-length vectors that encode relevant sequence information~\citeauthor{tayebi2021robust}. Common approaches include one-hot encoding~\citeauthor{kuzmin2020machine}, frequency-based encoding~\citeauthor{ali2021k}, and position-specific scoring matrices~\citeauthor{ali2022pwm2vec}. While these methods have provided valuable insights into protein analysis, they also come with certain drawbacks. One of the problems with these methods is that the important local and long-range interactions within the sequence may be overlooked~\citeauthor{wu2022sproberta}. Another challenge is the dimensionality of the embedding space~\citeauthor{yeung2023tree}. Protein sequences can be quite long, resulting in high-dimensional vectors. Furthermore, traditional embedding methods may struggle to capture fine-grained details and subtle variations in protein sequences~\citeauthor{ingraham2019generative}. They often treat each amino acid as independent, disregarding the context and spatial arrangements that are crucial for understanding protein structure and function~\citeauthor{ali2021k}.

Considering the drawbacks of traditional embedding methods, there is a need for a more advanced and efficient representation-learning approach that can overcome these limitations. Image-based representations, such as the Chaos Game Representation (CGR)~\citeauthor{jeffrey1990chaos} approach utilized in this study, offer a promising alternative by preserving sequential information, capturing spatial relationships, and enabling a more comprehensive analysis of protein sequences. Using the image-based representation also opens up the whole domain of deep learning for vision to be applied directly on the protein-based images, which is not possible in the case of traditional vector embeddings as deep learning methods do not perform well on tabular data~\citeauthor{lochel2020deep}.

\subsection{Chaos Game Representation (CGR)}
The CGR works by applying recursive chaos game rules on the protein sequences to generate the images~\citeauthor{lochel2020deep}. In this method, a central seed point is established, and successive iterations are performed using a set of predefined rules. With each iteration, the seed point is displaced based on the specific amino acid encountered in the sequence. The resulting movement generates patterns that unfold into symmetrical and visually captivating kaleidoscopic images~\citeauthor{nair2009bio}. The choice to use the kaleidoscopic-based image generation using the Chaos Game Representation (CGR) method is justified by its ability to generate visually captivating images that exhibit symmetrical patterns. While other CGR methods exist, such as n-flakes~\citeauthor{lochel2020deep}, the kaleidoscopic approach offers a unique aesthetic appeal that enhances the visualization of protein sequences. See Figre~\ref{fig_kaleidoscopic_sample} for an example of a kaleidoscopic shape image generated using chaos game representation.


The kaleidoscopic shape images generated through CGR provide a visually engaging representation of the underlying protein sequences. The symmetrical patterns created by the recursive chaos game rules reflect the inherent symmetries and repetitive motifs within the protein sequences. This can facilitate the identification of structural and functional patterns that may be important for understanding protein properties.
Furthermore, kaleidoscopic images offer an intuitive and visually accessible representation that can aid in the interpretation and analysis of protein sequences. The symmetrical nature of the patterns can help highlight and emphasize important features or regions within the sequence, allowing for a more intuitive understanding of the sequence's structural and functional characteristics.
By utilizing the kaleidoscopic approach, this study harnesses the unique visual properties of the generated images to provide a novel and aesthetically appealing representation of protein sequences. This visual representation can enhance the exploration and analysis of protein data, potentially leading to new insights and discoveries in the field of bioinformatics.

Deep learning has emerged as a powerful tool for image classification tasks~\citeauthor{li2019deep}. In this paper, we leverage deep learning techniques to perform classification on the generated chaos images. We design and train deep learning models, such as convolutional neural networks (CNNs), to learn the intricate patterns and features present in the chaos images. By training these models on the training set and evaluating their performance on the validation set, we aim to achieve an accurate and reliable classification of the protein sequences based on their visual representations.

The combination of chaos image generation and deep learning classification opens up new avenues for protein analysis and bioinformatics research~\citeauthor{lochel2020deep}. The application of deep learning models to classify the chaos images allows us to explore the relationship between the visual patterns observed in the kaleidoscopic images and the assigned labels. This classification can potentially uncover meaningful associations between specific visual patterns and protein characteristics, such as functional domains, secondary structures, or evolutionary relationships.


This paper makes several key contributions to the field of protein analysis and classification using the Chaos Game Representation (CGR) approach. Our contributions can be summarized as follows:

\begin{enumerate}
    \item \textbf{Introducing the use of CGR for generating kaleidoscopic images of protein sequences:} We showcase the application of CGR in visualizing protein sequences by recursively applying chaos game rules. Our proposed method, called \textbf{D}eep Learning-Assisted \textbf{A}nalysis of Protei\textbf{N} Sequences Using  \textbf{C}haos \textbf{E}nhanced Kaleidoscopic Images (DANCE), generates visually captivating kaleidoscopic shape images that capture the structural and functional characteristics of proteins.
    \item \textbf{Demonstrating the effectiveness of DANCE images for protein sequence classification:} We explore the utilization of DANCE images as visual representations for protein sequence classification. By employing deep learning image classifiers on the DANCE images, and demonstrate their efficacy in accurately categorizing protein sequences based on the visual patterns.
    \item \textbf{Investigating the relationship between visual patterns in DANCE images and protein properties:} We analyze the relationship between the visual patterns observed in the DANCE images and the underlying protein properties. This exploration provides insights into how the kaleidoscopic shape reflects structural motifs, protein domains, secondary structures, and other relevant features.
    \item \textbf{Bridging the gap between visual representations and protein classification:} This paper addresses the gap in existing research by integrating CGR-based DANCE images with deep learning techniques for protein sequence classification. We demonstrate the synergy between visual representations and computational models, enhancing our understanding of protein sequences comprehensively and intuitively.
\end{enumerate}


\section{Related Work}~\label{sec_related_work}


Sparse encoding~\citeauthor{hirst1992prediction} uses a one-hot binary vector of length 20 to represent each amino acid in a protein sequence. However, this approach suffers from inefficiency and redundancy due to its high-dimensional and sparse nature. Amino Acid Composition~\citeauthor{matsuda2005novel} offers an alternative protein representation by considering the local compositions of amino acids and their twins. However, it does not consider the sequence order, limiting its effectiveness. Physicochemical Properties~\citeauthor{deber2001tm} incorporate the molecular components' physicochemical properties to predict protein structure and function. However, the challenge lies in determining effective encoding for unknown physicochemical properties involved in protein folding. Notably, these feature engineering-based methods are domain-specific and may lack generalizability across different data types.
The structural-based encoding methods include Quantitative Structure-Activity Relationship (QSAR)~\citeauthor{cherkasov2014qsar} and General Structure encoding~\citeauthor{cui2008computational}. QSAR utilizes chemical properties to describe the amino acids in a sequence, but it focuses solely on the molecules rather than encoding the entire residue. However, QSAR may be susceptible to false correlations resulting from experimental errors in biological data. On the other hand, General Structure encoding maps structural information (e.g., residue depth, 3D shape, secondary structure) of the protein sequence into a numerical representation. However, its performance is limited by the availability of known protein structures.

Protein visualization techniques have played a crucial role in understanding protein structure and function~\citeauthor{cournia2015membrane}. Traditional methods, such as space-filling models~\citeauthor{matthews2017high}, provide valuable insights into the three-dimensional (3D) structure of proteins. However, these techniques often struggle to capture the intricate details of protein sequences and their relationships~\citeauthor{itoh2009hybrid}.
The Chaos Game Representation (CGR) has emerged as a powerful tool for visualizing DNA and RNA sequences~\citeauthor{lochel2021chaos}. By recursively applying chaos game rules to generate fractal-like patterns, CGR enables the visualization of sequence properties and motifs~\citeauthor{thomas2023three}. However, its application in protein sequence analysis remains relatively unexplored.
 In recent years, deep learning techniques have revolutionized various domains, including image classification~\citeauthor{affonso2017deep}. Also been applied to protein sequence classification tasks~\citeauthor{ao2022biological}. Convolutional Neural Networks (CNNs) and Recurrent Neural Networks (RNNs) have shown promising results in extracting meaningful features from protein sequences and achieving high classification accuracy.

Despite the advancements in protein visualization~\citeauthor{colaert2009improved}, CGR~\citeauthor{lochel2020deep}, and deep learning-based classification~\citeauthor{senior2020improved}, there exists a significant gap in the literature regarding the application of CGR to generate the kaleidoscopic shape of protein sequences. Most existing research focuses on either 3D protein structure visualization or DNA/RNA sequence analysis using CGR~\citeauthor{thomas2023three,lochel2020deep}. The potential of kaleidoscopic representations for capturing complex patterns and relationships within protein sequences remains largely unexplored.

\section{Proposed Approach}~\label{sec_proposed_model}
Our proposed approach, DANCE, combines the Chaos Game Representation (CGR) with advanced deep learning techniques to classify protein sequences effectively. This innovative method harnesses the power of visual representation and neural networks to capture complex patterns in protein sequences, aiming for improved accuracy and robustness in classification tasks. 

The Chaos Game Representation (CGR) is a method originally designed for visualizing sequences in a two-dimensional space. In the context of protein sequences, CGR converts linear sequences of amino acids into a 2D image, where each amino acid is mapped to a specific coordinate based on a set of predefined rules. These rules associate each amino acid with specific coordinates in the image, allowing us to create a visually informative representation of the protein sequence. The final output of this mapping process is a 2D image where the spatial distribution of pixels represents the sequence of amino acids in the protein. This image captures both the sequence order and the amino acid composition, offering a rich visual representation of the protein's structure. 
Our proposed approach comprised several steps, which we will now discuss one by one.



\subsection{Assign numerical Coordinates To Amino Acids}
The first step is to assign fixed x-axis and y-axis coordinate values to each of the $20$ possible amino acids in protein sequences. Although this assignment of coordinate values could be random, the only criterion is that the values should be unique. Each amino acid must be assigned a unique pair of coordinates. This uniqueness is essential to ensure that each amino acid can be distinctly represented and identified in the CGR image, avoiding any ambiguity or overlap between different amino acids. The proper assignment of coordinates is crucial for the CGR process because it determines how the amino acids are represented in the final 2D image. Accurate and unique coordinate assignment allows for clear and effective visualization of protein sequences, capturing their compositional and sequential characteristics in a manner that can be analyzed by deep learning models for various classification tasks. The x- and y-axis values assigned to each amino acid are given in Table~\ref{tbl_aa_axis}.

\begin{minipage}{.40\textwidth}
    \centering
    \includegraphics[scale=0.2]{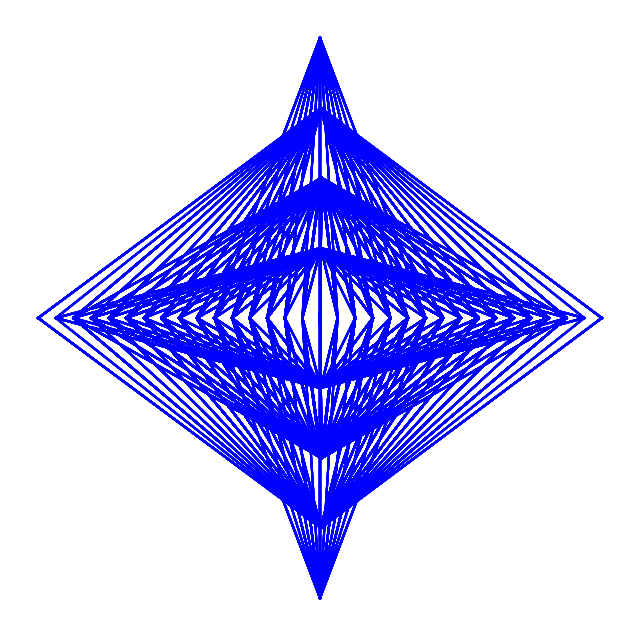}
    \captionof{figure}{A kaleidoscopic shape image generated using chaos game representation for a sample sequence ``ACQRSTAGTACGT".}
    \label{fig_kaleidoscopic_sample}
\end{minipage} \hfill
\hspace{0.01\linewidth}
\begin{minipage}{.480\textwidth}
    \centering
     \resizebox{0.8\textwidth}{!}{
    \begin{tabular}{ccc|ccc}
    \toprule
    Amino Acid & x-axis & y-axis & Amino Acid  & x-axis & y-axis \\
    \midrule \midrule
    A & 0.5 & 0.5 & M & 0.5 & 0.0 \\
    C & 1.0 & 0.5 & N & 0.25 & 0.5 \\
    D & 0.5 & 1.0 & P & 1.0 & 0.0 \\
    E & 0.0 & 0.5 & Q & 0.0 & 1.0 \\
    F & 1.0 & 1.0 & R & 0.5 & 0.25 \\
    G & 0.25 & 0.25 & S & 0.75 & 0.5 \\
    H & 0.75 & 0.25 & T & 0.5 & 0.75 \\
    I & 0.75 & 0.75 & V & 0.0 & 0.0 \\
    K & 0.25 & 0.75 & W & 1.0 & 0.25 \\
    L & 0.75 & 0.0 & Y & 1.0 & 0.75 \\
    \bottomrule
    \end{tabular}
    }
     \captionof{table}{Amino acids with corresponding \\ x- and y-axis values.}
    \label{tbl_aa_axis}
    \end{minipage}

\subsection{Recursively Generating DANCE Images}
The pseudocode to generate the Kaleidoscope shape images is given in Algorithm~\ref{algo_kaleidoscope}. This method takes a protein sequence as an input along with the recursion depth, initial position of the central seed point, initial angle of rotation, and scale factor for the replication. It recursively calls itself, keeps updating coordinate values, adding coordinates in the plot, and reducing the depth. When depth reaches $\leq 0$, the algorithm terminates (i.e. stopping criteria met) and the resultant plot is the final DANCE-based image for the given protein sequence. The variables depth, initial position (pos), angle, and scale are the hyperparameters, whole values are tuned using a standard validation set approach. The initial optimal values selected for the depth, initial position (pos), angle, and scale are $4$, (0, 0), 0, and $10$, respectively. After the recursive process terminates, we get the DANCE (Kaleidoscope shape) image (see Figure~\ref{fig_kaleidoscopic_sample} for an example). Once the Kaleidoscope shape image is generated, it is used as input for deep learning-based classifiers. The deep learning models analyze these images to classify the protein sequences, leveraging the visual patterns created by the CGR method to extract meaningful features for accurate classification. Figure~\ref{fig_kaleidoscopic_sample} illustrates a sample Kaleidoscope shape image generated using this method. The image showcases the intricate patterns that result from the recursive plotting of amino acid coordinates, demonstrating the effectiveness of the CGR technique in visualizing protein sequences and providing a unique and detailed representation of protein sequences, facilitating enhanced analysis and classification through deep learning models.






            

            
            
            
        
            
            
            
        

\begin{algorithm}[h!]
\caption{Generate Kaleidoscope (DANCE)}
\label{algo_kaleidoscope}
\begin{algorithmic}[1]
\scriptsize
\STATE \textbf{Input:} {Set $\mathcal{M}$ of ($m$-mer) minimizers on alphabet $\Sigma$}
\STATE \textbf{Output:} {ViralVectors based embedding $V$}
  
\STATE{GenKaleidoscope}($seq, depth, pos, angle, scale$)
    \IF{$depth \leq 0$}
        \STATE \textbf{return}
    \ENDIF
    \STATE $x, y \gets pos$
    \STATE $dx \gets scale \cdot \cos(angle)$
    \STATE $dy \gets scale \cdot \sin(angle)$
    
    \FOR{$AminoAcid$ \textbf{in} $seq$}
        \STATE $x, y \gets x + dx, y + dy$
        \STATE $cx, cy \gets \textsc{CoordinateRule}(AminoAcid)$ \COMMENT{from Table~\ref{tbl_aa_axis}}
        
        \STATE \text{plt.plot}([$x, cx$], [$y, cy$], \text{color}=$color$)
        \STATE \text{plt.plot}([$x, cx$], [$y, -cy$], \text{color}=$color$)
        \STATE \text{plt.plot}([$-x, cx$], [$-y, cy$], \text{color}=$color$)
        \STATE \text{plt.plot}([$-x, cx$], [$-y, -cy$], \text{color}=$color$)
        
        \STATE \textsc{GenKaleidoscope}$(seq, depth-1, (x, y), angle, scale)$
        \STATE \textsc{GenKaleidoscope}$(seq, depth-1, (x, -y), angle, scale)$
        \STATE \textsc{GenKaleidoscope}$(seq, depth-1, (-x, y), angle, scale)$
        \STATE \textsc{GenKaleidoscope}$(seq, depth-1, (-x, -y), angle, scale)$
        
        \STATE $depth \gets depth - 1$
    \ENDFOR
\end{algorithmic}
\end{algorithm}


  
    
        
        
        

\section{Experimental Setup}~\label{sec_experimental_setup}
This section presents details regarding the dataset used and the evaluation metrics employed in the experiments. The experiments were performed on a computer system equipped with an Intel(R) Core i5 processor, 32 GB of memory, and a 64-bit Windows 10 operating system. The models were implemented using the Python programming language. For the sake of reproducibility, we have made our preprocessed data and code available online~\footnote{The preprocessed data and code can be accessed in the published version of this work.}.

For assessing the effectiveness of the deep learning models, we measure several performance metrics, including average accuracy, precision, recall, F1 (weighted), F1 (macro), ROC-AUC, and training runtime. In the case of multi-class classification, we adopt the one-vs-rest approach to utilize binary classification-based evaluation metrics. This approach enables us to evaluate the model's performance across multiple classes. By using these metrics, we ensure a thorough evaluation of our deep learning models, addressing various aspects of performance from accuracy and error rates to computational efficiency. This comprehensive assessment helps in fine-tuning the models and making informed decisions about their deployment and application.

\subsection{Dataset Statistics}
The TCR sequence data used in this study was obtained from TCRdb, a comprehensive database for T-cell receptor sequences known for its powerful search function~\citeauthor{chen2021tcrdb}. 
In this study, our focus was on identifying and extracting data related to the five most prevalent types of cancer-based on their incidence rates. 
We extracted a total of $14205$ TCR sequences for four different types of cancers. 
We use the following target labels with the number of sequences: HeadNeck(5230), Ovarian(583), Pancreatic(2887), Retroperitoneal(5505).



\subsection{Feature Engineering Baselines}
In addition to the Chaos method~\citeauthor{lochel2020deep}, which serves as the state-of-the-art (SOTA) approach for comparison, we incorporate two numerical feature vector-based sequence embedding generation methods as baselines. The following sections provide detailed descriptions of these baselines.

\subsubsection{One Hot Encoding (OHE)~\citeauthor{kuzmin2020machine}}
OHE (One-Hot Encoding) is used to transform a sequence into a numerical representation. It creates a binary feature vector for each character in the sequence, and these binary vectors are then concatenated to represent the entire sequence. While OHE is a simple and intuitive method, the resulting vectors tend to be highly sparse, leading to challenges related to the curse of dimensionality.

\subsubsection{Wasserstein Distance Guided Representation Learning (WDGRL)~\citeauthor{shen2018wasserstein}}
This is an unsupervised domain adaptation technique that aims to transform high-dimensional vectors into low-dimensional representations. This approach utilizes neural networks to determine the Wasserstein distance (WD) between the encoded distributions of the source and target data. By optimizing the feature extractor network and minimizing the estimated WD, WDGRL obtains effective representations of the input data features. WDGRL operates on the feature vectors generated by the OHE method.

\subsubsection{Efficient Kernel~\citeauthor{ali2024molecular}}
Authors in~\citeauthor{ali2024molecular} propose a kernel-based method for molecular sequence classification, addressing challenges in detecting diseases using molecular data. The approach involves creating a kernel matrix using normalized pairwise $k$-mer distances, optimized via the Sinkhorn-Knopp algorithm, followed by kernel PCA to reduce dimensionality. We use this method with the logistic regression classifier (i.e. a commonly used classifier in the literature) as a baseline for cancer prediction.

\subsection{Classification Models}
To perform the classification of TCRs with respect to their cancer activity type we are employing two types of deep learning (DL) models, vision models \& tabular models.

The vision models consist of a set of DL classifiers that are applicable to the image dataset, and they are used to classify the TCR images generated by our proposed approach and the Chaos baseline. 
This set has $4$ custom convolution neural network (CNN) models along $2$ pre-trained classifiers. The custom classifiers are known as $1$-Layer CNN, $2$-Layer CNN, $3$-Layer CNN \& $4$-Layer CNN. Their names indicate the number of hidden Block layers present in them. For instance, in $4$-Layer CNN $4$ Block layers exist and a Block layer has a Convolution layer followed by a
ReLu activation function and a Max-Pool layer with a kernel size of 5x5 and stride of 2x2. In each of the custom models, the final layer comprises $2$ fully connected layers with the ReLu activation function and Softmax classification layer. These custom CNN classifiers illustrate the impact of increasing the number of layers in a classifier on the performance of the classifier. Moreover, the impact of transfer learning is observed by using the pre-trained models for the TCR classification task. We employ two pre-trained models, VGG-19~\citeauthor{Simonyan15} and RESNET-50~\citeauthor{he2016deep}, as both of them are very popular image classifiers. 
Furthermore, the 80-20\% train-test split is used for training the vision models based on stratified sampling. This sampling technique is known to preserve the proportions between the classes. The input images are of size $380 \times 380$. The training hyper-parameters used are $0.003$ learning rate, $64$ batch size, $10$ epochs, and ADAM optimizer chosen after fine-tuning the models. Additionally, the negative log-likelihood (NLL)~\citeauthor{yao2019negative} loss function is used as a training loss function because it’s known to be a cross-entropy loss function for multi-class problems.

The tabular CNN classifiers take vector data as input and these models are applied to the vectors generated from the feature-engineering-based baselines (OHE \& WDGRL). The tab CNN set contains $3$-Layer Tab CNN \& $4$-Layer Tab CNN model. Their names imply the number of hidden linear layers in them, like the $4$-Layer Tab CNN model has $4$ hidden fully connected layers. In both models, the hidden layers are followed by a final classification linear layer. Their training hyper-parameters are $0.003$ learning rate, $64$ batch size, $10$ epochs, ADAM optimizer, and NLL loss function. They also follow the 80-20\% train-test split in the training. Moreover, the WDGRL technique generates the vectors of dimension $10$, while OHE uses a zero padding strategy to make its vectors the same length. 

\section{Results and Discussion}~\label{sec_results_discussion}
This section deals with the classification results of TCRs based on their cancer activity type using various DL classifiers. The results are summarized in Table~\ref{tbl_cancer_data_results}.

\paragraph{\textbf{Comparision with feature-engineering-based baselines}}
The results illustrate that the feature-engineering-based baselines (OHE \& WDGRL) achieve lower performance using the tabular CNN models compared to our image-based method (DANCE) for all the evaluation metrics except the train run time. 
We can also observe that DANCE outperforms the efficient kernel method for all evaluation metrics, showing that the image-based approach captures the underlying sequence patterns more effectively than kernel-based embeddings. This suggests that transforming sequences into an image format allows deep learning models to better leverage spatial relationships and local dependencies within the data, leading to superior predictive performance. Additionally, DANCE's ability to outperform kernel-based methods highlights the advantage of using convolutional architectures for sequence classification, as they excel at recognizing complex structures in visual representations, which are often missed by traditional vector-based or kernel methods.

\paragraph{\textbf{Comparision with image-based baseline}}
We can observe that our method (DANCE) is outperforming the image-based baseline (Chaos) for all the evaluation metrics. This indicates that the images generated by DANCE are more informative in terms of classification performance than the images created by Chaos. 
Moreover, DANCE perform better for almost all the evaluation parameters corresponding to the $3$-Layer CNN model, along with the $1$-Layer CNN model also yielded optimal values for accuracy, recall, and AUC ROX scores. We can notice that increasing the number of layers to $3$ layers is increasing the performance for most of the metrics, while more than $3$ layers are demonstrating a decreased performance. One reason for that could be the gradient vanishing issue. As our dataset is not large, a higher number of layers in the model can cause the gradient to vanish, hence hindering the learning capacity of the model. 

Furthermore, we investigated transfer learning for doing TCR classification using the pre-trained RESNET-50 and VGG-19 models. The results illustrated that DANCE clearly performs better than the pre-trained models. A reason for that could be that the RESNET-50 and VGG-19 models are trained originally on different types of image data, so they are unable to generalize well to the DANCE-based images.

\begin{table*}[h!]
    \centering
    \caption{The TCR classification results for different models and algorithms. The best values are shown in bold.}
    \resizebox{0.9\textwidth}{!}{
    \begin{tabular}{cp{2.4cm}p{0.9cm}p{0.9cm}p{1.1cm}p{1.2cm}p{1.2cm}p{1.1cm}|p{1.4cm}}
    \toprule
        \multirow{2}{*}{DL Model} & \multirow{2}{*}{Method}  & \multirow{2}{*}{Acc. $\uparrow$} & \multirow{2}{*}{Prec. $\uparrow$} & \multirow{2}{*}{Recall $\uparrow$} & \multirow{2}{1.2cm}{F1 (Weig.) $\uparrow$} & \multirow{2}{1.4cm}{F1 (Macro) $\uparrow$} & \multirow{2}{1.2cm}{ROC AUC $\uparrow$} & Train Time (hrs.) $\downarrow$ \\
        \midrule \midrule
        - &
         Efficient Kernel~\citeauthor{ali2024molecular} & 0.386 & 0.149 & 0.386 & 0.215 & 0.139 & 0.500 & 1.207 \\  
         \midrule
         \multirow{2}{1.5cm}{3-Layer Tab CNN}  &
         OHE~\citeauthor{kuzmin2020machine} &  0.388 & 0.291 & 0.388 & 0.321 & 0.211 & 0.491 & 0.249 \\ 
         & WDGRL~\citeauthor{shen2018wasserstein} & 0.436 & 0.339 & 0.436 & 0.358 & 0.236 & 0.510 & \textbf{0.070} \\ 
         \midrule
         \multirow{2}{1.5cm}{4-Layer Tab CNN}  &
         OHE~\citeauthor{kuzmin2020machine} &  0.371 & 0.286 & 0.371 & 0.288 & 0.192 & 0.489 & 0.330 \\ 
         & WDGRL~\citeauthor{shen2018wasserstein} & 0.435 & 0.384 & 0.435 & 0.355 & 0.236 & 0.500 & 0.074 \\ 
         \midrule
        \multirow{2}{1.5cm}{1-Layer CNN} 
         & Chaos &  0.343 & 0.330 & 0.343 & 0.335 & 0.246 & 0.498 & 4.983 \\ 
         & DANCE (Ours) &  \textbf{0.478} & 0.440 & \textbf{0.478} & 0.312 & 0.278 & \textbf{0.635} & 3.099 \\ 
         \midrule
         \multirow{2}{1.5cm}{2-Layer CNN} 
         & Chaos &  0.381 & 0.285 & 0.381 & 0.215 & 0.140 & 0.499 & 5.183 \\ 
         & DANCE (Ours) & 0.460 & 0.407 & 0.460 & 0.394 & 0.264 & 0.544 & 3.101 \\ 
         \midrule
         \multirow{2}{1.5cm}{3-Layer CNN} 
         & Chaos &  0.379 & 0.143 & 0.379 & 0.208 & 0.137 & 0.500 & 6.156 \\ 
         & DANCE (Ours) &  \textbf{0.478} & \textbf{0.451} & \textbf{0.478} & \textbf{0.430} & \textbf{0.299} & 0.559 & 3.186 \\ 
         \midrule
         \multirow{2}{1.5cm}{4-Layer CNN} 
         & Chaos & 0.381 & 0.145 & 0.381 & 0.210 & 0.138 & 0.500 & 5.566 \\ 
         & DANCE (Ours) & 0.457 & 0.341 & 0.457 & 0.385 & 0.255 & 0.542 & 3.105 \\ 
         \midrule
         \multirow{2}{1.5cm}{PreTrained RESNET50} 
         & Chaos & 0.379 & 0.143 & 0.379 & 0.208 & 0.137 & 0.489 & 7.600 \\ 
         & DANCE (Ours) &  0.459 & 0.343 & 0.459 & 0.393 & 0.261 & 0.501 & 8.152 \\ 
         \midrule
         \multirow{2}{1.5cm}{PreTrained VGG-19} 
         & Chaos &  0.379 & 0.143 & 0.379 & 0.208 & 0.137 & 0.488 & 16.420 \\ 
         & DANCE (Ours) & 0.430 & 0.320 & 0.430 & 0.366 & 0.243 & 0.500 & 15.643 \\ 
        \bottomrule
    \end{tabular}
    }
    \label{tbl_cancer_data_results}
\end{table*}

\section{Conclusion}\label{sec_conclusion}
In conclusion, this study presents the DANCE (Deep Learning-Assisted Analysis of Protein Sequences Using Chaos Enhanced Kaleidoscopic Images) approach, which combines Chaos Game Representation (CGR) with deep learning classification to address the challenges in analyzing T-cell protein sequences. By generating kaleidoscopic images using CGR, DANCE offers a visually captivating representation that preserves essential details and captures the structural and functional characteristics of protein sequences. The effectiveness of DANCE images for protein sequence classification is demonstrated through the utilization of deep learning models. Additionally, the study investigates the relationship between the visual patterns observed in DANCE images and protein properties, providing insights into structural motifs, protein domains, secondary structures, and other relevant features. By bridging the gap between visual representations and protein classification, this research contributes to the field of protein analysis and bioinformatics, offering new possibilities for a comprehensive and intuitive understanding of protein sequences. Future work includes evaluation of DANCE on other biological datasets such as coronavirus spike sequences and Zika virus sequences etc. Using more advanced deep learning models, such as Transformers for image classification is another exciting future extension.

\nocite*{}



\section*{Author Contribution Statement}

\textbf{SA, MP:} Conceptualization. \textbf{TM, SA:}
Methodology. \textbf{PC, TM:} Software. \textbf{PC, SA:}
Validation. \textbf{PC, SA:} Formal Analysis. \textbf{All:}
Investigation. \textbf{All:} Resources. \textbf{MP:} Data
Curation. \textbf{All:} Writing - Original Draft. \textbf{All:}
Writing - Review \& Editing. \textbf{PC, SA:}
Supervision. \textbf{MP:}
Administration. \textbf{SA, MP:} Funding Acquisition.

\section*{Conflict of Interest}

The authors declare no conflict of interest.

\section*{Funding Statement}

Research supported by an MBD Fellowship for SA, PC, and a Georgia State University Computer Science start-up grant for MP.  

\bibliographystyle{plainnat-revised}
\bibliography{references}

\begin{thebibliography}{62}
\providecommand{\natexlab}[1]{#1}
\providecommand{\url}[1]{\texttt{#1}}
\expandafter\ifx\csname urlstyle\endcsname\relax
  \providecommand{\doi}[1]{doi: #1}\else
  \providecommand{\doi}{doi: \begingroup \urlstyle{rm}\Url}\fi

\bibitem[Affonso et~al.(2017)Affonso, Rossi, et~al.]{affonso2017deep}
Affonso, C., Rossi, A. L.~D., et~al.
\newblock Deep learning for biological image classification.
\newblock \emph{Expert systems with applications}, 85:\penalty0 114--122, 2017.

\bibitem[Ali and Patterson(2021)]{ali2021spike2vec}
Ali, S. and Patterson, M.
\newblock {Spike2vec}: An efficient and scalable embedding approach for
  covid-19 spike sequences.
\newblock In \emph{IEEE International Conference on Big Data (Big Data)}, pages
  1533--1540, 2021.

\bibitem[Ali et~al.(2021)Ali, Sahoo, Ullah, Zelikovskiy, Patterson, and
  Khan]{ali2021k}
Ali, S., Sahoo, B., Ullah, N., Zelikovskiy, A., Patterson, M., and Khan, I.
\newblock A k-mer based approach for {SARS-CoV-2} variant identification.
\newblock In \emph{International Symposium on Bioinformatics Research and
  Applications}, pages 153--164, 2021.

\bibitem[Ali et~al.(2022)Ali, Bello, Chourasia, Punathil, Zhou, and
  Patterson]{ali2022pwm2vec}
Ali, S., Bello, B., Chourasia, P., Punathil, R.~T., Zhou, Y., and Patterson, M.
\newblock {PWM2Vec}: An efficient embedding approach for viral host
  specification from coronavirus spike sequences.
\newblock \emph{Biology}, 11\penalty0 (3):\penalty0 418, 2022.

\bibitem[Ali et~al.(2023)Ali, Murad, and Patterson]{ali2023pssm2vec}
Ali, S., Murad, T., and Patterson, M.
\newblock Pssm2vec: A compact alignment-free embedding approach for coronavirus
  spike sequence classification.
\newblock In \emph{Neural Information Processing: 29th International
  Conference, ICONIP 2022, Virtual Event, November 22--26, 2022, Proceedings,
  Part VII}, pages 420--432. Springer, 2023.

\bibitem[Ali et~al.(2024)Ali, Ali, Murad, Mansoor, and
  Patterson]{ali2024molecular}
Ali, S., Ali, T.~E., Murad, T., Mansoor, H., and Patterson, M.
\newblock Molecular sequence classification using efficient kernel based
  embedding.
\newblock \emph{Information Sciences}, 679:\penalty0 121100, 2024.

\bibitem[Ao et~al.(2022)Ao, Jiao, Wang, Yu, and Zou]{ao2022biological}
Ao, C., Jiao, S., Wang, Y., Yu, L., and Zou, Q.
\newblock Biological sequence classification: A review on data and general
  methods.
\newblock \emph{Research}, 2022:\penalty0 0011, 2022.

\bibitem[Barnsley(2012)]{barnsley2012fractals}
Barnsley, M.~F.
\newblock Fractals everywhere: New edition, 2012.

\bibitem[Bourne et~al.(2022)Bourne, Draizen, and Mura]{bourne2022curse}
Bourne, P.~E., Draizen, E.~J., and Mura, C.
\newblock The curse of the protein ribbon diagram.
\newblock \emph{PLoS biology}, 20\penalty0 (12):\penalty0 e3001901, 2022.

\bibitem[Campbell et~al.(2022)Campbell, Gifford, Singer, Hill, O’toole,
  Rambaut, Hampson, and Brunker]{campbell2022making}
Campbell, K., Gifford, R.~J., Singer, J., Hill, V., O’toole, A., Rambaut, A.,
  Hampson, K., and Brunker, K.
\newblock Making genomic surveillance deliver: A lineage classification and
  nomenclature system to inform rabies elimination.
\newblock \emph{PLoS Pathogens}, 18\penalty0 (5):\penalty0 e1010023, 2022.

\bibitem[Chen et~al.(2021{\natexlab{a}})Chen, Zha, Zhu, Ning, and
  Cui]{chen2021hydrogen}
Chen, C., Zha, Y., Zhu, D., Ning, K., and Cui, X.
\newblock Hydrogen bonds meet self-attention: all you need for protein
  structure embedding.
\newblock In \emph{2021 IEEE International Conference on Bioinformatics and
  Biomedicine (BIBM)}, pages 12--17. IEEE, 2021{\natexlab{a}}.

\bibitem[Chen et~al.(2021{\natexlab{b}})Chen, Yue, Lei, and Guo]{chen2021tcrdb}
Chen, S.-Y., Yue, T., Lei, Q., and Guo, A.-Y.
\newblock Tcrdb: a comprehensive database for t-cell receptor sequences with
  powerful search function.
\newblock \emph{Nucleic Acids Research}, 49\penalty0 (D1):\penalty0 D468--D474,
  2021{\natexlab{b}}.

\bibitem[Cherkasov et~al.(2014)]{cherkasov2014qsar}
Cherkasov, A. et~al.
\newblock Qsar modeling: where have you been? where are you going to?
\newblock \emph{Journal of medicinal chemistry}, 57\penalty0 (12):\penalty0
  4977--5010, 2014.

\bibitem[Chowdhury and Garai(2017)]{chowdhury2017review}
Chowdhury, B. and Garai, G.
\newblock A review on multiple sequence alignment from the perspective of
  genetic algorithm.
\newblock \emph{Genomics}, 109\penalty0 (5-6):\penalty0 419--431, 2017.

\bibitem[Colaert et~al.(2009)Colaert, Helsens, Martens,
  et~al.]{colaert2009improved}
Colaert, N., Helsens, K., Martens, L., et~al.
\newblock Improved visualization of protein consensus sequences by icelogo.
\newblock \emph{Nature methods}, 6\penalty0 (11):\penalty0 786--787, 2009.

\bibitem[Cournia et~al.(2015)Cournia, Allen, Andricioaei,
  et~al.]{cournia2015membrane}
Cournia, Z., Allen, T.~W., Andricioaei, I., et~al.
\newblock Membrane protein structure, function, and dynamics: a perspective
  from experiments and theory.
\newblock \emph{The Journal of membrane biology}, 248:\penalty0 611--640, 2015.

\bibitem[Cui et~al.(2008)Cui, Liu, Puett, and Xu]{cui2008computational}
Cui, J., Liu, Q., Puett, D., and Xu, Y.
\newblock Computational prediction of human proteins that can be secreted into
  the bloodstream.
\newblock \emph{Bioinformatics}, 24\penalty0 (20):\penalty0 2370--2375, 2008.

\bibitem[Deber et~al.(2001)Deber, Wang, et~al.]{deber2001tm}
Deber, C.~M., Wang, C., et~al.
\newblock Tm finder: a prediction program for transmembrane protein segments
  using a combination of hydrophobicity and nonpolar phase helicity scales.
\newblock \emph{Protein Science}, 10\penalty0 (1):\penalty0 212--219, 2001.

\bibitem[Dosovitskiy et~al.(2020)Dosovitskiy, Beyer, Kolesnikov, Weissenborn,
  Zhai, Unterthiner, Dehghani, Minderer, Heigold, Gelly,
  et~al.]{dosovitskiy2020image}
Dosovitskiy, A., Beyer, L., Kolesnikov, A., Weissenborn, D., Zhai, X.,
  Unterthiner, T., Dehghani, M., Minderer, M., Heigold, G., Gelly, S., et~al.
\newblock An image is worth 16x16 words: Transformers for image recognition at
  scale.
\newblock \emph{arXiv preprint arXiv:2010.11929}, 2020.

\bibitem[Du et~al.(2020)Du, He, Li, and Uversky]{du2020deepadd}
Du, Z., He, Y., Li, J., and Uversky, V.~N.
\newblock Deepadd: protein function prediction from k-mer embedding and
  additional features.
\newblock \emph{Computational Biology and Chemistry}, 89, 2020.

\bibitem[Eisenberg(1984)]{eisenberg1984three}
Eisenberg, D.
\newblock Three-dimensional structure of membrane and surface proteins.
\newblock \emph{Annual review of biochemistry}, 53\penalty0 (1):\penalty0
  595--623, 1984.

\bibitem[Farhan et~al.(2017)Farhan, Tariq, Zaman, Shabbir, and
  Khan]{farhan2017efficient}
Farhan, M., Tariq, J., Zaman, A., Shabbir, M., and Khan, I.
\newblock Efficient approximation algorithms for strings kernel based sequence
  classification.
\newblock In \emph{Advances in neural information processing systems
  (NeurIPS)}, pages 6935--6945. ., 2017.

\bibitem[Gohil et~al.(2021)Gohil, Iorgulescu, Braun, Keskin, and
  Livak]{gohil2021applying}
Gohil, S.~H., Iorgulescu, J.~B., Braun, D.~A., Keskin, D.~B., and Livak, K.~J.
\newblock Applying high-dimensional single-cell technologies to the analysis of
  cancer immunotherapy.
\newblock \emph{Nature Reviews Clinical Oncology}, 18\penalty0 (4):\penalty0
  244--256, 2021.

\bibitem[Greenwade(1993)]{greenwade93}
Greenwade, G.~D.
\newblock The {C}omprehensive {T}ex {A}rchive {N}etwork ({CTAN}).
\newblock \emph{TUGBoat}, 14\penalty0 (3):\penalty0 342--351, 1993.

\bibitem[He et~al.(2016)He, Zhang, Ren, and Sun]{he2016deep}
He, K., Zhang, X., Ren, S., and Sun, J.
\newblock Deep residual learning for image recognition.
\newblock In \emph{IEEE conference on computer vision and pattern recognition},
  pages 770--778, 2016.

\bibitem[Hirst and Sternberg(1992)]{hirst1992prediction}
Hirst, J.~D. and Sternberg, M.~J.
\newblock Prediction of structural and functional features of protein and
  nucleic acid sequences by artificial neural networks.
\newblock \emph{Biochemistry}, 31\penalty0 (32):\penalty0 7211--7218, 1992.

\bibitem[Hopp and Woods(1981)]{hopp1981prediction}
Hopp, T.~P. and Woods, K.~R.
\newblock Prediction of protein antigenic determinants from amino acid
  sequences.
\newblock \emph{Proceedings of the National Academy of Sciences}, 78\penalty0
  (6):\penalty0 3824--3828, 1981.

\bibitem[Hou et~al.(2016)Hou, Wang, Lu, Xie, Cui, Chen, Du, Dai, and
  Diao]{hou2016analysis}
Hou, X., Wang, M., Lu, C., Xie, Q., Cui, G., Chen, J., Du, Y., Dai, Y., and
  Diao, H.
\newblock Analysis of the repertoire features of tcr beta chain cdr3 in human
  by high-throughput sequencing.
\newblock \emph{Cellular Physiology and Biochemistry}, 39\penalty0
  (2):\penalty0 651--667, 2016.

\bibitem[Ingraham et~al.(2019)Ingraham, Garg, Barzilay, and
  Jaakkola]{ingraham2019generative}
Ingraham, J., Garg, V., Barzilay, R., and Jaakkola, T.
\newblock Generative models for graph-based protein design.
\newblock \emph{Advances in neural information processing systems}, 32, 2019.

\bibitem[Itoh et~al.(2009)Itoh, Muelder, Ma, and Sese]{itoh2009hybrid}
Itoh, T., Muelder, C., Ma, K.-L., and Sese, J.
\newblock A hybrid space-filling and force-directed layout method for
  visualizing multiple-category graphs.
\newblock In \emph{2009 IEEE Pacific Visualization Symposium}, pages 121--128.
  IEEE, 2009.

\bibitem[Jeffrey(1990)]{jeffrey1990chaos}
Jeffrey, H.~J.
\newblock Chaos game representation of gene structure.
\newblock \emph{Nucleic acids research}, 18\penalty0 (8):\penalty0 2163--2170,
  1990.

\bibitem[Kabsch and Sander(1983)]{kabsch1983dictionary}
Kabsch, W. and Sander, C.
\newblock Dictionary of protein secondary structure: pattern recognition of
  hydrogen-bonded and geometrical features.
\newblock \emph{Biopolymers: Original Research on Biomolecules}, 22\penalty0
  (12):\penalty0 2577--2637, 1983.

\bibitem[Kuzmin et~al.(2020)Kuzmin, Adeniyi, DaSouza~Jr, Lim, Nguyen, Molina,
  Xiong, Weber, and Harrison]{kuzmin2020machine}
Kuzmin, K., Adeniyi, A.~E., DaSouza~Jr, A.~K., Lim, D., Nguyen, H., Molina,
  N.~R., Xiong, L., Weber, I.~T., and Harrison, R.~W.
\newblock Machine learning methods accurately predict host specificity of
  coronaviruses based on spike sequences alone.
\newblock \emph{Biochemical and Biophysical Research Communications},
  533\penalty0 (3):\penalty0 553--558, 2020.

\bibitem[Kyte and Doolittle(1982)]{kyte1982simple}
Kyte, J. and Doolittle, R.~F.
\newblock A simple method for displaying the hydropathic character of a
  protein.
\newblock \emph{Journal of molecular biology}, 157\penalty0 (1):\penalty0
  105--132, 1982.

\bibitem[Li et~al.(2020)Li, Yuan, Tian, Meng, and Liu]{li2020t}
Li, N., Yuan, J., Tian, W., Meng, L., and Liu, Y.
\newblock T-cell receptor repertoire analysis for the diagnosis and treatment
  of solid tumor: a methodology and clinical applications.
\newblock \emph{Cancer Communications}, 40\penalty0 (10):\penalty0 473--483,
  2020.

\bibitem[Li et~al.(2019)Li, Song, Fang, Chen, Ghamisi, and
  Benediktsson]{li2019deep}
Li, S., Song, W., Fang, L., Chen, Y., Ghamisi, P., and Benediktsson, J.~A.
\newblock Deep learning for hyperspectral image classification: An overview.
\newblock \emph{IEEE Transactions on Geoscience and Remote Sensing},
  57\penalty0 (9):\penalty0 6690--6709, 2019.

\bibitem[L{\"o}chel et~al.(2020)L{\"o}chel, Eger, Sperlea, and
  Heider]{lochel2020deep}
L{\"o}chel, H.~F., Eger, D., Sperlea, T., and Heider, D.
\newblock Deep learning on chaos game representation for proteins.
\newblock \emph{Bioinformatics}, 36\penalty0 (1):\penalty0 272--279, 2020.

\bibitem[L{\"o}chel and Heider(2021)]{lochel2021chaos}
L{\"o}chel, H.~F. and Heider, D.
\newblock Chaos game representation and its applications in bioinformatics.
\newblock \emph{Computational and Structural Biotechnology Journal},
  19:\penalty0 6263--6271, 2021.

\bibitem[Ma et~al.(2020)Ma, Yu, Tang, et~al.]{ma2020phylogenetic}
Ma, Y., Yu, Z., Tang, R., et~al.
\newblock Phylogenetic analysis of hiv-1 genomes based on the position-weighted
  k-mers method.
\newblock \emph{Entropy}, 22\penalty0 (2):\penalty0 255, 2020.

\bibitem[MacCallum and Tieleman(2011)]{maccallum2011hydrophobicity}
MacCallum, J.~L. and Tieleman, D.~P.
\newblock Hydrophobicity scales: a thermodynamic looking glass into
  lipid--protein interactions.
\newblock \emph{Trends in biochemical sciences}, 36\penalty0 (12):\penalty0
  653--662, 2011.

\bibitem[Matsuda et~al.(2005)Matsuda, Vert, Saigo, Ueda, Toh, and
  Akutsu]{matsuda2005novel}
Matsuda, S., Vert, J.-P., Saigo, H., Ueda, N., Toh, H., and Akutsu, T.
\newblock A novel representation of protein sequences for prediction of
  subcellular location using support vector machines.
\newblock \emph{Protein Science}, 14\penalty0 (11):\penalty0 2804--2813, 2005.

\bibitem[Matthews et~al.(2017)Matthews, Easdon, Kitao, Hayward, and
  Laycock]{matthews2017high}
Matthews, N., Easdon, R., Kitao, A., Hayward, S., and Laycock, S.
\newblock High quality rendering of protein dynamics in space filling mode.
\newblock \emph{Journal of Molecular Graphics and Modelling}, 78:\penalty0
  158--167, 2017.

\bibitem[Murad et~al.(2023)Murad, Ali, Chourasia, Mansoor, and
  Patterson]{murad2023circular}
Murad, T., Ali, S., Chourasia, P., Mansoor, H., and Patterson, M.
\newblock Circular arc length-based kernel matrix for protein sequence
  classification.
\newblock In \emph{2023 IEEE International Conference on Big Data (BigData)},
  pages 1429--1437. IEEE, 2023.

\bibitem[Nair et~al.(2009)Nair, Nair, et~al.]{nair2009bio}
Nair, A.~S., Nair, V.~V., et~al.
\newblock Bio-sequence signatures using chaos game representation.
\newblock \emph{Bioinformatics: applications in life and environmental
  sciences}, pages 62--76, 2009.

\bibitem[O'Shea and Nash(2015)]{o2015introduction}
O'Shea, K. and Nash, R.
\newblock An introduction to convolutional neural networks.
\newblock \emph{arXiv preprint arXiv:1511.08458}, 2015.

\bibitem[Otter et~al.(2020)Otter, Medina, and Kalita]{otter2020survey}
Otter, D.~W., Medina, J.~R., and Kalita, J.~K.
\newblock A survey of the usages of deep learning for natural language
  processing.
\newblock \emph{IEEE transactions on neural networks and learning systems},
  32\penalty0 (2):\penalty0 604--624, 2020.

\bibitem[{Protein Subcellular Localization}(2022)]{ProtLoc_website_url}
{Protein Subcellular Localization}.
\newblock
  \url{https://www.kaggle.com/datasets/lzyacht/proteinsubcellularlocalization},
  2022.
\newblock [Online; accessed 10-October-2022].

\bibitem[{RABV-GLUE Website}()]{RABV_GLUE_website_url}
{RABV-GLUE Website}.
\newblock \url{http://rabv-glue.cvr.gla.ac.uk/#/home}.

\bibitem[Senior et~al.(2020)Senior, Evans, Jumper, et~al.]{senior2020improved}
Senior, A.~W., Evans, R., Jumper, J., et~al.
\newblock Improved protein structure prediction using potentials from deep
  learning.
\newblock \emph{Nature}, 577\penalty0 (7792):\penalty0 706--710, 2020.

\bibitem[Shen et~al.(2018)Shen, Qu, Zhang, and Yu]{shen2018wasserstein}
Shen, J., Qu, Y., Zhang, W., and Yu, Y.
\newblock Wasserstein distance guided representation learning for domain
  adaptation.
\newblock In \emph{AAAI conference on artificial intelligence}, 2018.

\bibitem[Simonyan and Zisserman(2015)]{Simonyan15}
Simonyan, K. and Zisserman, A.
\newblock Very deep convolutional networks for large-scale image recognition.
\newblock In \emph{International Conference on Learning Representations}, 2015.

\bibitem[Tan and Le(2019)]{tan2019efficientnet}
Tan, M. and Le, Q.
\newblock Efficientnet: Rethinking model scaling for convolutional neural
  networks.
\newblock In \emph{International conference on machine learning}, pages
  6105--6114. PMLR, 2019.

\bibitem[Tayebi et~al.(2021)Tayebi, Ali, and Patterson]{tayebi2021robust}
Tayebi, Z., Ali, S., and Patterson, M.
\newblock Robust representation and efficient feature selection allows for
  effective clustering of {SARS-CoV-2} variants.
\newblock \emph{Algorithms}, 14\penalty0 (12), 2021.

\bibitem[Thomas(2023)]{thomas2023three}
Thomas, A.
\newblock Three dimensional chaos game representation of protein sequences.
\newblock \emph{arXiv preprint arXiv:2303.09683}, 2023.

\bibitem[Tzanov(2015)]{tzanov2015strictly}
Tzanov, V.
\newblock Strictly self-similar fractals composed of star-polygons that are
  attractors of iterated function systems.
\newblock \emph{arXiv preprint arXiv:1502.01384}, 2015.

\bibitem[Van~der Maaten and Hinton(2008)]{van2008visualizing}
Van~der Maaten, L. and Hinton, G.
\newblock Visualizing data using t-sne.
\newblock \emph{Journal of machine learning research}, 9\penalty0 (11), 2008.

\bibitem[Whisstock and Lesk(2003)]{whisstock2003prediction}
Whisstock, J.~C. and Lesk, A.~M.
\newblock Prediction of protein function from protein sequence and structure.
\newblock \emph{Quarterly reviews of biophysics}, 36\penalty0 (3):\penalty0
  307--340, 2003.

\bibitem[Wu et~al.(2022)Wu, Yin, Zhu, et~al.]{wu2022sproberta}
Wu, L., Yin, C., Zhu, J., et~al.
\newblock Sproberta: protein embedding learning with local fragment modeling.
\newblock \emph{Briefings in Bioinformatics}, 23\penalty0 (6), 2022.

\bibitem[Yao et~al.(2019)]{yao2019negative}
Yao et~al.
\newblock Negative log likelihood ratio loss for deep neural network
  classification.
\newblock In \emph{Proceedings of the Future Technologies Conference}, pages
  276--282. Springer, 2019.

\bibitem[Yeung et~al.(2023)Yeung, Zhou, Mathew, et~al.]{yeung2023tree}
Yeung, W., Zhou, Z., Mathew, L., et~al.
\newblock Tree visualizations of protein sequence embedding space enable
  improved functional clustering of diverse protein superfamilies.
\newblock \emph{Briefings in Bioinformatics}, 24\penalty0 (1):\penalty0
  bbac619, 2023.

\bibitem[Zhang et~al.(2020)Zhang, Bi, Wang, Zeng, Liao, and Chen]{9070714}
Zhang, J., Bi, C., Wang, Y., Zeng, T., Liao, B., and Chen, L.
\newblock Efficient mining closed k-mers from dna and protein sequences.
\newblock In \emph{2020 IEEE International Conference on Big Data and Smart
  Computing (BigComp)}, pages 342--349, 2020.
\newblock \doi{10.1109/BigComp48618.2020.00-51}.

\bibitem[{Zubair Hassan}(2022)]{EfficeintNet}
{Zubair Hassan}.
\newblock 3 pre-trained image classification models.
\newblock \url{https://www.folio3.ai/blog/image-classification-models/}, 2022.

\end{thebibliography}

\end{document}